# The Case for Universal Basic Computing Power

Yue Zhu1,2*
1 Tongji University School of Law
2 Shanghai Collaborative Innovation Center of AI Governance
* Corresponding author. Email: yue_zhu@tongji.edu.cn

We are currently observing two divergent trends in AI development. On one side, there is an exponential growth in the amount of training data, the number of model parameters, and the computational requirements of the underlying models. [1-3] This escalating need for scalability is outpacing the capabilities of individuals and most enterprises. This trajectory suggests a concerning future where AI could become increasingly centralized and exclusive. Conversely, there is still hope for AI to remain open and widespread, as evidenced by initiatives like the open-sourcing of foundational models such as LLaMA 2 and Claude 2, and techniques for model distillation, compression, and efficient deployment. As of now, an individual with adequate AI literacy can feasibly train and use a foundation model on a personal device.

To foster a more inclusive AI future, we should minimize the centralization trend while bolstering open AI initiatives. This entails not only acknowledging the contributions of the open-source community but also broadening the beneficiaries of these initiatives beyond a handful of tech-savvy individuals. This group likely represents a tiny fraction of privileged individuals. To address this disparity, we introduce the Universal Basic Computing Power (UBCP) initiative.

The UBCP initiative ensures global, free access to a set amount of computing power specifically for AI research and development (R&D). This initiative comprises three key elements. First, UBCP must be cost-free, with its usage limited to AI R&D and minimal additional conditions. Second, UBCP should continually incorporate the state-of-the-art AI advancements, including efficiently distilled, compressed, and deployed training data, foundational models, benchmarks, and governance tools. Lastly, it's essential for UBCP to be universally accessible, ensuring convenience for all users. We urge major stakeholders in AI development—large platforms, open-source contributors, and policymakers—to prioritize the UBCP initiative.

We now explain in detail the what, why, who, and how of UBCP.

The foundational premise of UBCP is its cost-free nature, inspired by the Universal Basic Income (UBI) initiative. Much like UBI aims to provide everyone with an unconditional financial foundation, UBCP seeks to offer computational resources to all without discrimination. It is crucial to note that access to UBCP should not be contingent upon one's technological literacy or knowledge. In fact, UBCP aims to bridge the gap created by the very lack of such knowledge or literacy.

The second core tenet of UBCP is its commitment to staying updated with AI's latest advancements, encompassing data, models, benchmarks, and governance tools. Merely providing computing power is not the end goal; instead, the emphasis is on enhancing AI literacy and facilitating smoother adaptation to the AI era for everyone.

Taking cues from industry-established low-code AI platforms can guide UBCP's design. Integral to UBCP should be extensive datasets, robust foundation models, and universally accepted benchmarks. By adopting a user-friendly design, those utilizing UBCP can effortlessly integrate these modules into AI applications. Furthermore, enriching open-source platforms with detailed datasheets and model cards paves the way for comprehensive AI understanding, steering it towards societal benefit. [4-5]

The third fundamental principle of UBCP is its universal accessibility. Ensuring consistent experiences for users worldwide, regardless of their location, is a paramount challenge. Key considerations include making UBCP fully functional on mobile devices, given the prevalence of mobile technology, especially in underserved communities. The user interface of UBCP should adhere to accessibility standards, ensuring an inclusive design. Special attention should be paid to catering to diverse cognitive needs, such as those of children and the elderly. Techniques like visualization, animation, and gamification can aid in demystifying complex AI concepts. Additionally, thorough localization, including the translation of technical terms, is crucial for global adoption.

The rationale behind the UBCP initiative can be understood at two levels: the broader human perspective and the specific interests of AI stakeholders. Many of the arguments championing UBI can analogously apply to UBCP. Summarized, these benefits revolve around three central themes: empowerment, individualization, and autonomy. [6] Firstly, UBCP provides everyone, regardless of background, the means to cultivate AI literacy and stay abreast of rapid technological shifts. Secondly, it allows individuals to tailor AI solutions to their unique needs, promoting inclusivity and reducing the risk of technological monopolies imposing a one-size-fits-all approach. Lastly, UBCP equips each person with the tools and space, reminiscent of Virginia Woolf's concept of a personal room, to explore, innovate, and harness the potential of AI.

It is also worth emphasizing that UBCP sidesteps some of the pitfalls commonly associated with UBI. Principal concerns regarding UBI include its potential to induce inflation and diminish work incentives. In contrast, distributing computing power, as proposed by UBCP, would not lead to inflation. If anything, an increase in the democratization of computation might be celebrated by the AI community. Furthermore, rather than discouraging engagement, the very nature of UBCP — effective only when actively utilized — serves as a motivational tool, encouraging individuals to immerse themselves in technology and drive innovation.

It is imperative to recognize that while UBCP promises universal benefits, its successful implementation hinges on the support and involvement of key AI stakeholders. Taking a long-term perspective, we are convinced that all significant stakeholders will benefit from the UBCP initiative. When a technology scales to encompass billions, as witnessed in the evolution of the Internet, the resulting value creation can eclipse short-term trade-offs. To address immediate concerns, we highlight the following: For dominant platforms with foundational models, pursuing open-source policies can be mutually beneficial. Models available through UBCP could potentially pave the way for the integration of more robust commercial models. For the open-source community, there lies an opportunity to significantly shape UBCP's evolution, thereby positioning themselves as pivotal actors in AI's future trajectory. Lastly, for governmental bodies, an UBCP rich in governance tools offers a compelling solution, effectively aligning with ethical and normative AI expectations.

Implementing and managing the UBCP will be a formidable endeavor, given the global scale of its intended reach. Indeed, everyone worldwide stands as a stakeholder in this initiative. However, the initial formation and success of the UBCP will heavily rely on three primary entities: prominent platforms, the open-source community, and policymakers and legislators. Prominent platforms, especially those at the forefront of cloud computing services, are uniquely positioned to spearhead UBCP's early stages. Their established infrastructure can serve as UBCP's backbone, and their vast computing resources can handle the extensive demand UBCP will generate. Moreover, while ensuring the UBCP remains functional and accessible, they can also introduce individual users to more sophisticated commercial AI solutions where applicable. If such integration fosters empowerment, addresses specific needs, and furthers autonomy, it is worth pursuing.

The open-source community will play a pivotal role in shaping the specifics of the UBCP, from data selection to model choice, benchmarks, and governance tools. When we emphasize "state of the art" repeatedly, we envision the adoption of the most relevant datasets, models, benchmarks, and governance tools that resonate with current trends on open-source platforms. Representative organizations or platforms from the open-source sector should be integral to the UBCP's foundation and operation, guiding AI's future trajectory from a technical standpoint.

Policymakers and legislators across countries, including major players like the European Union, the United States, and China, hold indispensable roles both individually and collaboratively. While fairness and autonomy are well-established AI governance principles, their practical implementation remains a challenge. [7-8] The UBCP offers a tangible opportunity to address this. For instance, both Article 4a of the EU's Artificial Intelligence Law, currently under intense negotiation, and Articles 4 and 8 of China's Model Law on Artificial Intelligence (Version 1.1), emphasize AI's principles of fairness, non-discrimination, and autonomy. The UBCP aligns perfectly with these principles. On a national level, countries should encourage, through their policies and laws, the establishment and operation of UBCPs that are unconditional, free, and open to all, or at the very least, they should not deter such initiatives. Collaboratively, countries should consider a globally coordinated UBCP through international agreements. Given AI's global reach and the cooperative nature of its development, despite the unequal distribution of its benefits and costs currently, promoting its open and democratic growth worldwide is essential. Thus, making UBCP a global endeavor is paramount.

As we stand on the brink of a transformative era driven by AI, our collective responsibility is to ensure that this technology benefits humanity at large. The UBCP emerges as a beacon of hope, promising equal access and fair distribution of AI's potential benefits. However, its success hinges not only on its conceptual appeal but on the practical steps we take today. Stakeholders, ranging from governments to industry leaders and the global citizenry, must come together to advocate for, fund, and implement UBCP initiatives. Only through concerted efforts can we realize the full democratic potential of AI, ensuring that it becomes a tool for progress and inclusivity rather than one of exclusion and disparity.